%% file: neurips_2025.tex
\documentclass{article}

\usepackage[final]{neurips_2025}

\usepackage[utf8]{inputenc} 
\usepackage[T1]{fontenc}    
\usepackage{hyperref}       
\usepackage{url}            
\usepackage{booktabs}       
\usepackage{amsfonts}       
\usepackage{nicefrac}       
\usepackage{microtype}      
\usepackage[table]{xcolor}         

\usepackage{graphicx}
\usepackage{multirow}
\usepackage[normalem]{ulem}
\usepackage{fonttable}
\usepackage{booktabs}
\usepackage{caption}
\usepackage{graphicx}
\usepackage{soul}
\usepackage{wrapfig}
\usepackage{booktabs}
\usepackage{caption}
\usepackage{lipsum} 
\usepackage{amsmath}

\title{GLVD: Guided Learned Vertex Descent}

\author{%
  Pol Caselles Rico \\
  Institut de Robotica i Informatica Industrial, CSIC-UPC \\
  Crisalix SA \\
  Barcelona, Spain \\
  \texttt{pcaselles22@gmail.com} \\
   \And
  Francesc Moreno Noguer\thanks{This work was conducted independently and does not relate to the author’s position at Amazon} \\
  Amazon \\
  Barcelona, Spain \\
  \texttt{cescmore@amazon.es} \\
}

\input{commands}

\begin{document}

\maketitle

\begin{figure}[h]
    \centering
    \includegraphics[width=1.\textwidth]{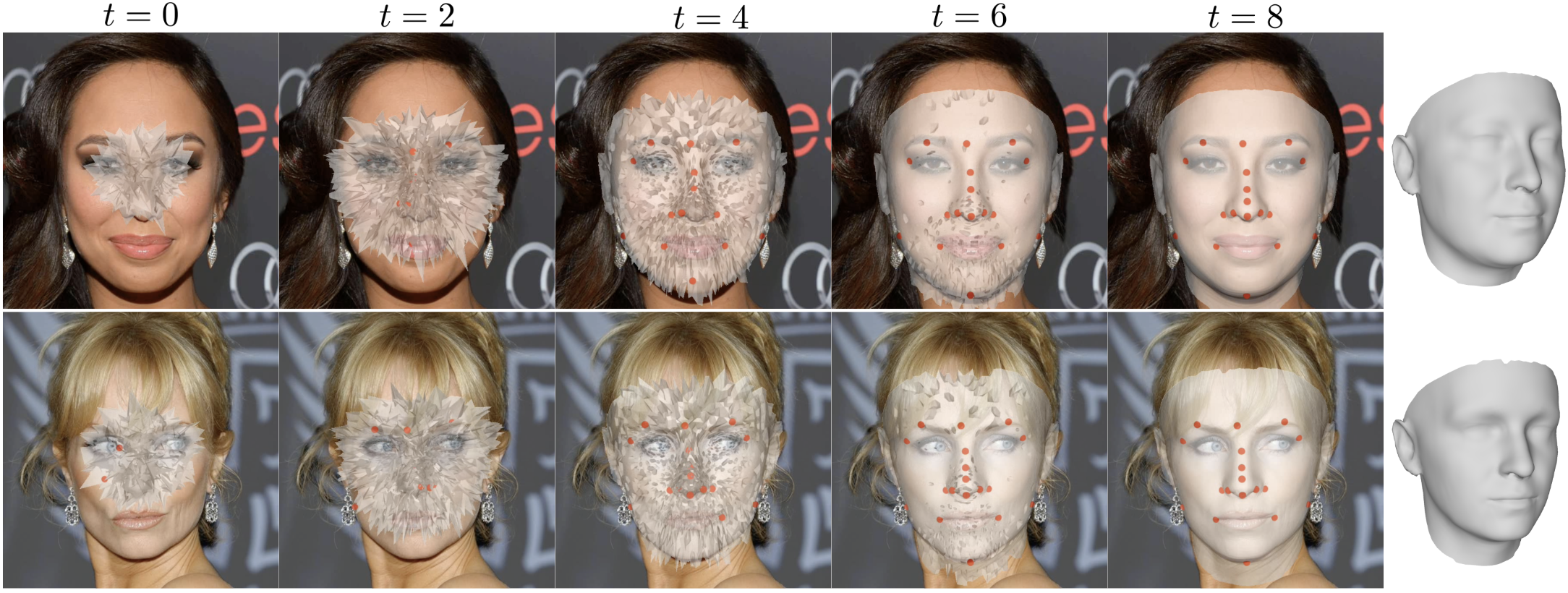}
    
    \caption{ Qualitative results for two in-the-wild subjects reconstructed using \method. }
    \label{fig:teaser}
\end{figure}

\input{Sections/01_abstract}
\input{Sections/02_introduction}

\input{Sections/03_related_work}

\input{Sections/04_method}

\input{Sections/05_experiments}


\input{Sections/06_conclusion}

\section{Acknowledgments}
This work has been supported by the project GRAVATAR PID2023-151184OB-I00 funded by MCIU/AEI/10.13039/501100011033 and by ERDF, UE.

\newpage
\small\bibliography{egbib}
\bibliographystyle{plain}

\newpage

\appendix
\newpage
\input{Sections/07_appendix}

\end{document}

%% file: commands.tex
\newcommand{\stkout}[1]{\ifmmode\text{\sout{\ensuremath{#1}}}\else\sout{#1}\fi}

\newcommand{\method}{GLVD }



%% file: Sections/01_abstract.tex
\begin{abstract}

Existing 3D face modeling methods usually depend on 3D Morphable Models, which inherently constrain the representation capacity to fixed shape priors. Optimization-based approaches offer high-quality reconstructions but tend to be computationally expensive. In this work, we introduce \textbf{GLVD}, a hybrid method for 3D face reconstruction from few-shot images that extends Learned Vertex Descent (LVD)~\cite{corona2022learned} by integrating per-vertex neural field optimization with global structural guidance from dynamically predicted 3D keypoints. By incorporating relative spatial encoding, GLVD iteratively refines mesh vertices without requiring dense 3D supervision. This enables expressive and adaptable geometry reconstruction while maintaining computational efficiency. GLVD achieves state-of-the-art performance in single-view settings and remains highly competitive in multi-view scenarios, all while substantially reducing inference time.

\end{abstract}

%% file: Sections/02_introduction.tex
\section{Introduction}

High-fidelity 3D face modeling from images is a long-standing challenge in the computer vision community, with broad impact across applications such as Virtual Reality, Augmented Reality, healthcare, entertainment, and security. Reconstructing an accurate and coherent digital human representation from a few input images is a highly ill-posed task -particularly in uncontrolled environments— often requiring geometry-aware methods guided by strong prior assumptions. Adding to this challenge is the scarcity of abundant, high-quality 3D training data captured under such unconstrained conditions, which limits the generalization ability of existing models in diverse real-world scenarios.

Statistical priors based on parametric 3D Morphable Models~\citep{bai2020deep, dou2018multi, Moreno_pami2013, ramon2019multi, richardson20163d, richardson2017learning, tewari2017mofa, tran2018extreme, tuan2017regressing, wu2019mvf, feng2021learning, MICA:ECCV2022, zhu2023facescape, wang2022faceverse, lei2023hierarchical} have become the standard approach for few-shot 3D face reconstruction. By encoding facial geometry using a low-dimensional set of parameters, 3DMMs provide a robust and efficient framework, particularly effective in scenarios with limited or single-view image input. However, their effectiveness is hindered by two key limitations: a bias toward the mean shape~\citep{sengupta2020synthetic}, and the inherently constrained expressiveness of parametric models. These models typically operate within fixed low-dimensional subspaces, making it difficult to capture fine-grained details or adapt to out-of-distribution variations.

Model-free representations using voxels~\cite{jackson2017large}, meshes, point clouds, or Gaussian splatting~\citep{kerbl3Dgaussians} offer greater flexibility and high reconstruction accuracy, but they face scalability and resolution trade-offs due to memory and topology constraints.  Neural fields address these challenges by encoding geometry and appearance as continuous functions via neural networks. These methods can reconstruct detailed surfaces from images without requiring 3D supervision, but they typically depend on multi-view inputs and suffer from high inference costs~\citep{martin2021nerf, barron2023zip}. Recent works ~\citep{mueller2022instant, neus2, canela2023instantavatar} have significant progress in reducing computational overhead. However, converting such representations into well-structured, topologically consistent meshes suitable for animation or rendering often necessitates additional post-processing, commonly involving template fitting. 

Optimization-based approaches produce accurate and detailed results through iterative refinement~\cite{martin2021nerf, neus2, canela2023instantavatar, ramon2021h3d, Caselles_2023_SIRA, caselles2025implicit}, while feed-forward methods~\citep{saito2019pifu, saito2020pifuhd, he2020geo, guo2023rafare} offer faster inference at the cost of robustness and accuracy, especially under out-of-distribution conditions~\citep{marin2024nicp}. More recently, Learned Vertex Descent (LVD)~\citep{corona2022learned} introduced a hybrid strategy that uses pixel-aligned image features to guide iterative template fitting. Despite its effectiveness, LVD relies on large-scale training data with posed images and corresponding 3D geometry, and it lacks explicit global structure—predicting vertex trajectories independently and depending on the image encoder for implicit coherence.

To overcome these limitations, we propose leveraging a 3D face landmark estimator derived from a 2D image-based predictor to guide 3D shape refinement. We introduce \method, learning-based optimization approach that fuses local and global cues by combining per-vertex neural fields with dynamically predicted 3D keypoints. Each neural field predicts 3D displacements for its associated vertex based on local features sampled at its current position, while the keypoint ensemble provides global structural guidance that informs and regularizes the optimization process. Central to our method is a relative encoding scheme, where each vertex is transformed based on the current keypoint estimates, allowing the network to learn geometry-aware updates that are conditioned on the evolving global structure.

The combination of local neural fields and global keypoint-based guidance in \method enables more precise control and adaptive refinement of 3D facial geometry as shown in Figure~\ref{fig:teaser}. Leveraging this fusion, we conduct a comprehensive evaluation on both single-view and multi-view 3D face reconstruction benchmarks. Our approach achieves state-of-the-art performance in single-image reconstruction and remains competitive with optimization-based methods in multi-view scenarios, demonstrating its robustness, accuracy, and broad applicability.

%% file: Sections/03_related_work.tex
\section{Related work}

\textbf{3D Morphable Models (3DMM).} The use of 3D Morphable Models (3DMMs) has become the standard paradigm for reconstructing 3D facial geometry from images, particularly in single-view or few-shot scenarios. These statistical models~\cite{paysan20093d, FLAME:SiggraphAsia2017, HIFI3D} are widely adopted and mainly focus on the facial region. In single-image settings, several methods have demonstrated effective reconstruction performance~\cite{tewari2017mofa, feng2021learning, MICA:ECCV2022, wang2022faceverse, zhu2023facescape, lei2023hierarchical}.

Recent advancements in single-image 3D face reconstruction have explored both parametric and non-parametric strategies to improve accuracy, robustness, and detail preservation. 3DDFAv2~\citep{guo2020towards} proposes a regression-based approach combining a lightweight architecture with meta-joint optimization to achieve real-time performance while maintaining alignment accuracy. Building upon this, 3DDFAv3~\citep{wang20243d} introduces Part Re-projection Distance Loss, which leverages dense facial part segmentation as a strong geometric prior for guiding 3D reconstruction, especially under extreme expressions where landmarks are unreliable. SADRNet~\citep{ruan2021sadrnet} introduces a self-aligned dual-regression framework that disentangles pose-dependent and pose-independent features and fuses them through an occlusion-aware alignment strategy. HRN~\citep{lei2023hierarchical} proposes a hierarchical representation network that disentangles geometric components and incorporates high-frequency priors, enabling the reconstruction of fine facial details, such as wrinkles and skin texture, from in-the-wild images.


\textbf{Neural Fields for Face Reconstruction.} Neural fields have emerged as a leading approach for 3D reconstruction, offering continuous and high-fidelity representations of geometry and appearance. They have been successfully applied to full-head and facial modeling tasks using techniques such as volume rendering and surface priors~\cite{mildenhall2020nerf, niemeyer2020differentiable, park2021nerfies, ramon2021h3d, Caselles_2023_SIRA, caselles2025implicit, canela2023instantavatar}. Hybrid models combine parametric approaches such as 3DMMs with neural implicit functions to increase control and expressiveness. For instance, IMFace~\cite{zheng2022imface} and IMFace++~\citep{zheng2023imface++} introduce implicit displacement fields to refine a 3DMM geometry, and NeuFace~\citep{zheng2023neuface} proposed an approximated BRDF integration and a low-rank prior for human face rendering. In \citep{cao2022jiff}, authors combine geometry-aware features with image features that output a signed distance field. However, these approaches tends to collapse and generate artifacts.

When several input images are available, a line of research \citep{grassal2021neural, zheng2022IMavatar, zheng2023pointavatar, bharadwaj2023flare, zielonka2023instant, giebenhain2024mononphm} aims to obtain animatable full-head avatars from videos. Building on the recent success of Gaussian Splatting~\citep{kerbl3Dgaussians}, several works~\citep{luo2024splatface, dhamo2024headgas, qian2024gaussianavatars, tang2024gaf} have integrated this representation to improve rendering efficiency and visual fidelity. Combining them with 3DMMs has been explored in recent works, with methods such as HeadGAP~\citep{zheng2024headgap} and GPHM~\citep{xu2024gphm} that learns parametric head models using Gaussian-splatting-based models. While effective, these methods often rely on dense multi-view input. In contrast, model-free feed-forward approaches use pixel-aligned features~\cite{saito2019pifu} for faster inference from sparse views.

\begin{figure}[!t]
    \centering
    \includegraphics[width=1.\textwidth]{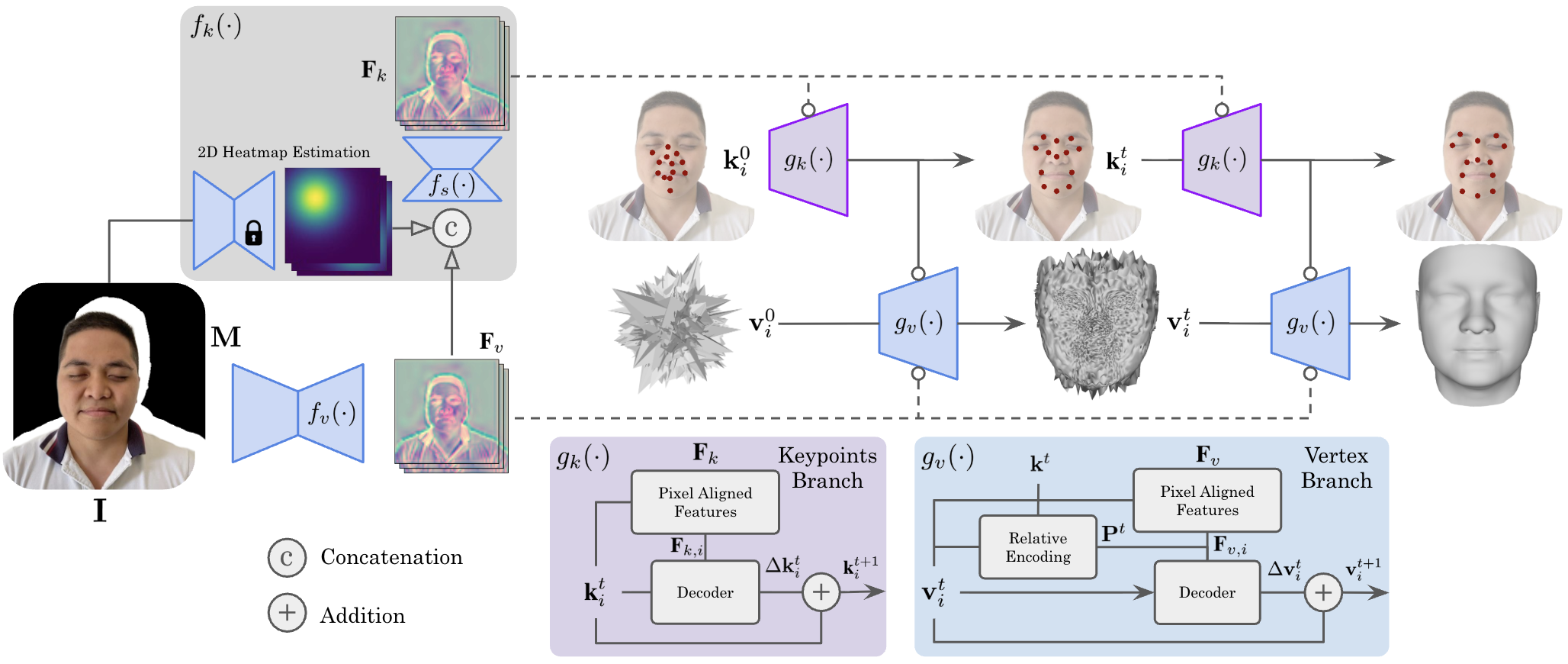}
    \caption{
    \textbf{Overview of \method.} Given one or more input images, each paired with a head mask and calibrated camera parameters, the method reconstructs a 3D face mesh through two branches. (1) The 3D Keypoint Branch predicts a set of facial keypoints by extracting localized image features and estimating their 3D displacements iteratively. (2) The 3D Vertex Branch refines the full-face geometry by leveraging these keypoints to encode relative spatial information for each surface vertex. This branch extracts pixel-aligned features and predicts vertex-wise displacements in an iterative optimization process. 
}
    \label{fig:method}
\end{figure}

Model-free methods leveraging pixel-aligned features have gained popularity for fast 3D reconstruction, as they avoid the need for test-time optimization~\cite{saito2019pifu, saito2020pifuhd, he2020geo, alldieck2022photorealistic, shao2022doublefield, corona2023structured, guo2023rafare}. PIFu~\cite{saito2019pifu} introduced pixel-aligned implicit functions, using 2D image features to predict 3D occupancy from single or multiple views, and Phorhum~\cite{alldieck2022photorealistic} extended this by employing signed distance fields for surface modeling. JIFF~\citep{cao2022jiff} proposed combining features from a face morphable model and pixel-aligned features. In contrast, using volumetric rendering, KeypointNeRF aggregates pixel-aligned features with a relative spatial encoder. More recently, LVD~\citep{corona2022learned} emerged as a learning-based
optimization approach that laverages pixel-aligned features to guide an iterative-based template fitting. While it acquires a good trade-off between computation requirements and accuracy for human mesh recovery, it remains unexplored for 3D face modeling. However, single feed-forward methods based on pixel-aligned features still lag behind optimization-based approaches in terms of reconstruction quality~\citep{caselles2025implicit}.

\textbf{Encoding Representations for 3D Reconstruction.} Previous approaches have explored various spatial encoding strategies to enhance learning. PVA~\citep{raj2021pva} and PortraitNeRF~\citep{gao2020portrait} utilize face-centric coordinate systems, while ARCH~\citep{huang2020arch} and ARCH++~\citep{he2021arch++} adopt canonical body coordinates. KeypointNeRF~\citep{Mihajlovic:ECCV2022} proposes to encode relative spatial 3D information in the form of depth via sparse 3D keypoints. In \citep{de2004sparse} the authors introduced a three-step pipeline of landmark selection, low-dimensional embedding via MDS, and distance-based triangulation to embed points. GLVD follows a similar idea by selecting identity-specific facial keypoints and encoding mesh vertices through their Euclidean distances to these keypoints within the learned space. In this work, we conduct a thorough investigation of spatial encoding and find that a simple yet effective encoding based on relative distances w.r.t 3D keypoints \citep{corona2021smplicit} yields effective results in combination with neural fields guided by pixel-aligned features. We adopt a canonical aligned space to stabilize training. In addition to achieving state-of-the-art results on 3D face reconstruction from as few as single input image, our approach can also be used beyond face modeling. 

\textbf{Mesh Recovery for body.} Several works~\citep{lin2021end, yoshiyasu2023deformable, ma20233d, lin2023one, zhang2023pymaf} focus on full-body mesh recovery and remain unexplored in the context of face modeling. These methods often operate in single-image settings~\citep{lin2021end, yoshiyasu2023deformable, ma20233d, lin2023one} or on video sequences~\citep{lei2024gart}. METRO \citep{lin2021end} and DeFormer \citep{yoshiyasu2023deformable} use transformers to jointly process mesh joints and vertices, differing in attention aggregation but both relying on global self-attention. METRO further conditions on a global image embedding, discarding pixel-aligned spatial detail. In contrast, 3D Virtual Markers \citep{ma20233d} predicts latent 3D markers and reconstructs the mesh as a linear combination, while One-Stage Mesh Recovery \citep{lin2023one} directly regresses SMPL parameters via a transformer. Both are constrained by the limited expressiveness of low-dimensional latent spaces. PyMAF-X \citep{zhang2023pymaf} introduces an iterative refinement approach that samples pixel-aligned features from prior vertex predictions, but does not leverage explicit landmark or keypoint constraints for structural supervision. GART \citep{lei2024gart} leverages skeletal priors and temporal video input for avatar reconstruction, whereas GLVD is purely image-driven and effective in both monocular and few-shot settings. Finally, \citep{ranjan2018generating} replaces PCA-based 3D face models with convolutional mesh autoencoders to learn shape priors, but focuses on latent mesh encoding rather than direct facial geometry estimation from RGB images as in GLVD

%% file: Sections/04_method.tex
\begin{figure}[!t]
    \centering
    \includegraphics[width=1.\textwidth]{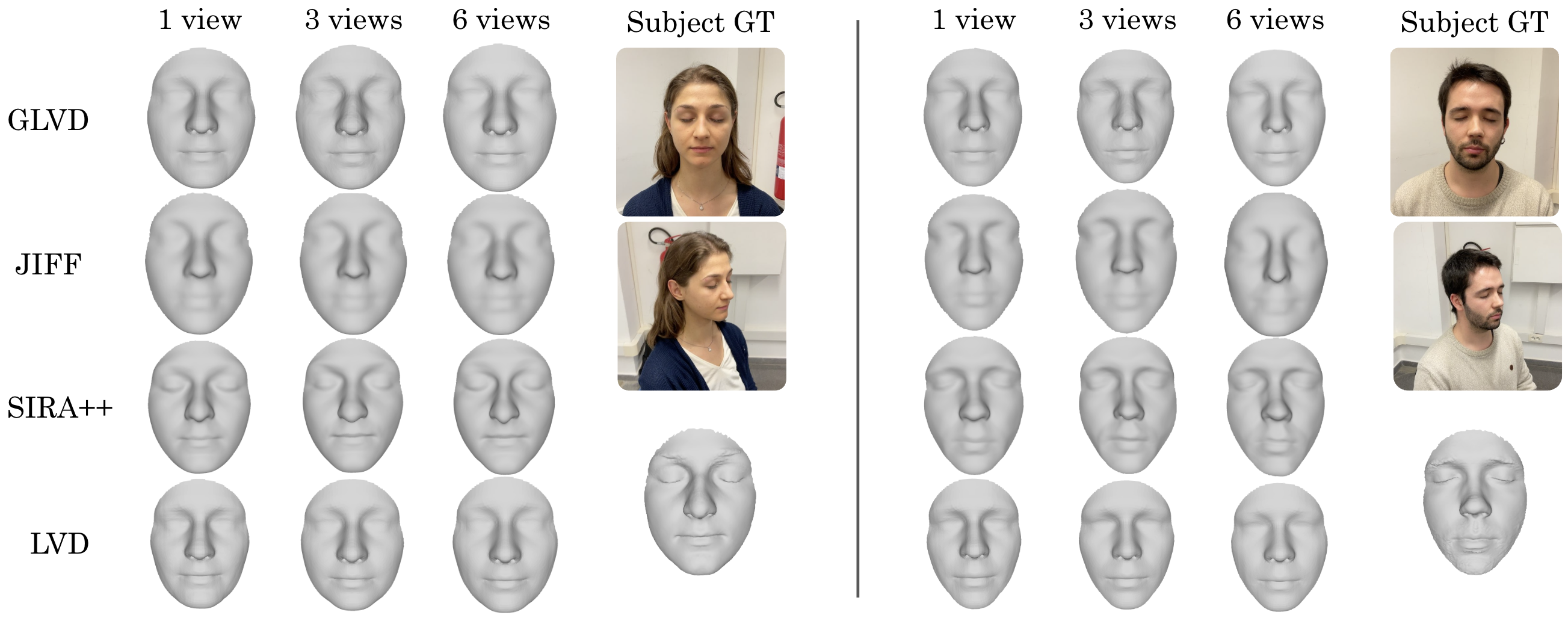}
    \caption{
    \textbf{Qualitative results on two subjects of the H3DS dataset}, for LVD~\citep{corona2022learned}, SIRA++\citep{caselles2025implicit}, JIFF~\citep{cao2022jiff}, and \method, with an increasing number of input views.
    }
    \label{fig:multiview}
\end{figure}

\section{Method}
\label{sec:method}

In this section, we present our method for 3D face reconstruction using learned keypoint guidance. We first review the Learned Vertex Descent (LVD) framework~\citep{corona2022learned}, then introduce our key architectural innovations. The section concludes with details on training and inference. An overview is shown in Figure~\ref{fig:method}, with implementation details in the supplementary material.

\subsection{Background: Learned Vertex Descent}
\label{sec:lvd_background}
Learned Vertex Descent (LVD)~\citep{corona2022learned} is an optimization-based method for 3D human shape reconstruction from single-view images or scans. While it has been applied to full-body and hand reconstruction, we explore its potential for 3D face modeling. The model learns a transformation  \( g(\cdot) \) which takes the current 3D vertex position at iteration \( t \) and the associated 2D image features \( \mathbf{f}_i \) for vertex  \( i \) as input,  and outputs a displacement vector  \( \Delta \mathbf{v}_i \):
\begin{equation}
    g : (\mathbf{v}_i^t, \mathbf{f}_i) \mapsto \Delta \mathbf{v}_i.
    \label{eq:main}
\end{equation}
This vector represents the correction needed to align the vertex with its ground truth position \( \hat{\mathbf{v}}_i \). The updated vertex position is then given by \( \mathbf{v}_i^{t+1} = \mathbf{v}_i^t + \Delta \mathbf{v}_i \). The reconstruction process involves iteratively applying this update to refine the 3D shape.

\subsection{Problem Definition}

Our goal is to recover a 3D face surface \( \mathcal{S} \) from a small set of input images \( \{\mathbf{I}_n\}_{n=1}^N \), where each image \( \mathbf{I}_n \) is paired with a head mask \( \mathbf{M}_n \) and calibrated camera parameters \( \mathbf{T}_n \). The surface \( \mathcal{S} \) is represented by a fixed topology of 7,225 vertices and 14,342 faces.

We aim to incorporate global 3D-aware guidance into the per-vertex optimization by leveraging a relative encoding based on the Euclidean distances between vertices and keypoints. As a result, we propose a two-stage architecture to address limited multi-view input images: (1) a 3D keypoint estimation module that defines spatial keypoints on the facial surface by estimating displacements, and (2) a vertex prediction module that encodes vertices relative to these keypoints to estimate vertex updates. Our formulation does not rely on a predefined parametric model or fixed joint sets, making it adaptable to arbitrary topologies. A sparse set of surface points is conveniently selected to act as ground-truth keypoints.

\begin{figure}[!t]
    \centering
    \includegraphics[width=1.\textwidth]{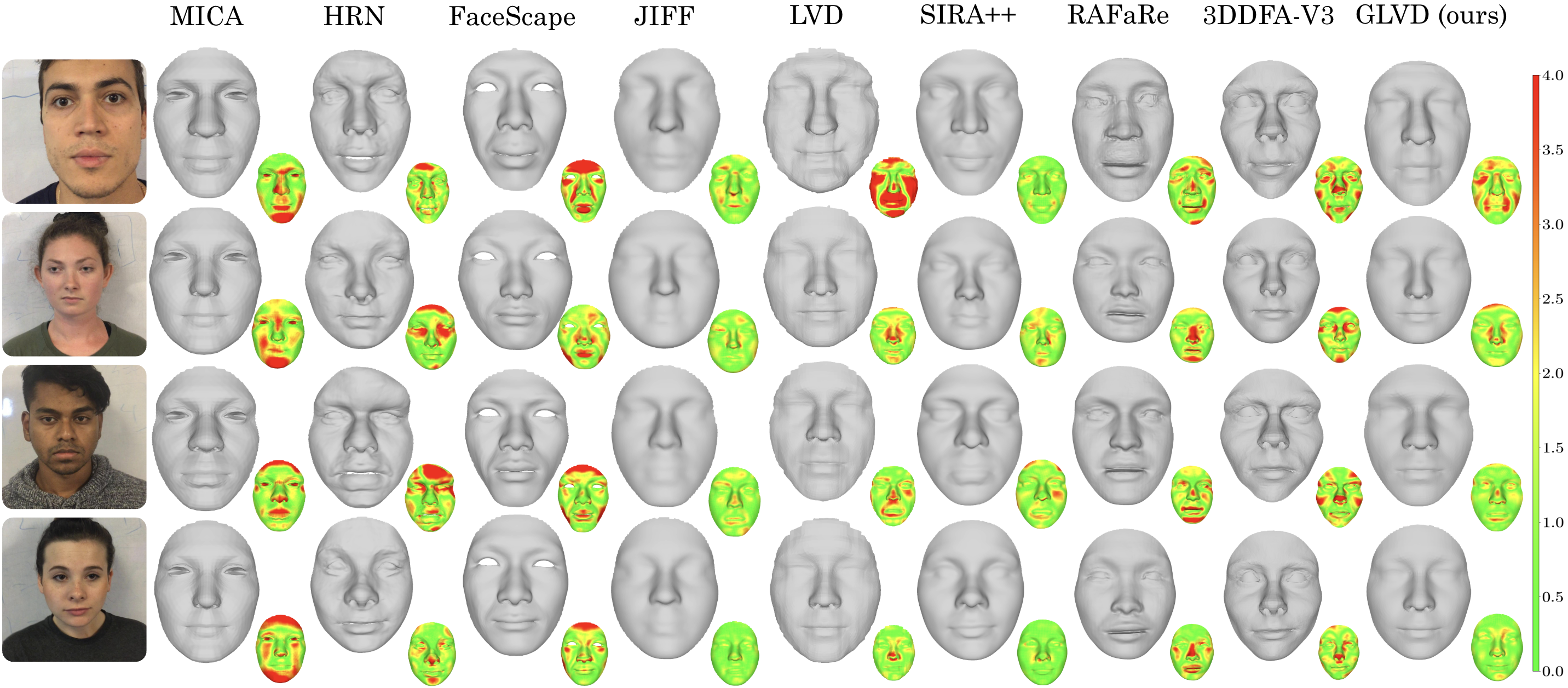}
    \caption{\textbf{Qualitative results on the 3DFAW dataset for a single input image.} Each 3D reconstructed face is accompanied by a heatmap, where reddish
    areas indicate larger errors in mm.
    }
    \label{fig:qualitative_1_view}
\end{figure}

\subsection{Learning-Based Keypoint Estimation}
\label{sec:method_keypoints}

Our goal is to establish a reference space to guide vertex optimization towards the target surface. To achieve this, we estimate a fixed set of \(K\) 3D keypoints  \( \{ \mathbf{k}_j^t \}_{j=1}^K \), where \( \mathbf{k}_j^t \in \mathbb{R}^3 \) represents the \(j\)-th keypoint on the target surface, at iteration \( t \). These keypoints are used for encoding the relative positions of query vertices \( \mathbf{v}_i \).

The 3D keypoint branch consists of two components: one responsible for extracting image features \( \mathbf{F}_k \), and another focused on learning 3D keypoint displacements \( \Delta\mathbf{k}_i \). To compute \( \mathbf{F}_k \), we first generate facial keypoint heatmaps using off-the-shelf HRNet~\citep{wang2020deep}, which are then concatenated with vertex image features \( \mathbf{F}_v \) obtained from the first stack of the Hourglass network \citep{newell2016stacked}. This combined feature map is subsequently refined using a single-stack Hourglass module \( f_s(\cdot) \):
\begin{center}
    \begin{equation}
        f_k : (\mathbf{I}, \mathbf{M}, \textbf{F}_{v}) \mapsto \textbf{F}_{k}.
    \end{equation}
\end{center}

The keypoints employed to guide the vertex branch do not necessarily coincide with the facial landmarks defined by HRNet. We adopt a strategy of predicting vertex displacements from local features, as this approach has been demonstrated to yield more accurate geometric detail~\citep{saito2019pifu, saito2020pifuhd, chibane2020implicit, corona2022learned}. To estimate the 3D keypoints, we implement \( g_k(\cdot) \) (Eq.~\ref{eq:main}) with a 3-layer MLP that takes as input the current estimate of keypoint \( \textbf{k}_{i}^t \) and its local F-dimensional local features \( \textbf{F}_{k,i} \) extracted at the projection of \( \textbf{k}_{i}^{t} \) on the image plane, and predicts the displacement \( \Delta\mathbf{k}_{i}^t \). In the first step, we begin by uniformly sampling  \(\mathbf{k}_i^0\) within a volume of size 2 centered at the origin.

\begin{figure}[!t]
    \centering
    \includegraphics[width=1.\textwidth]{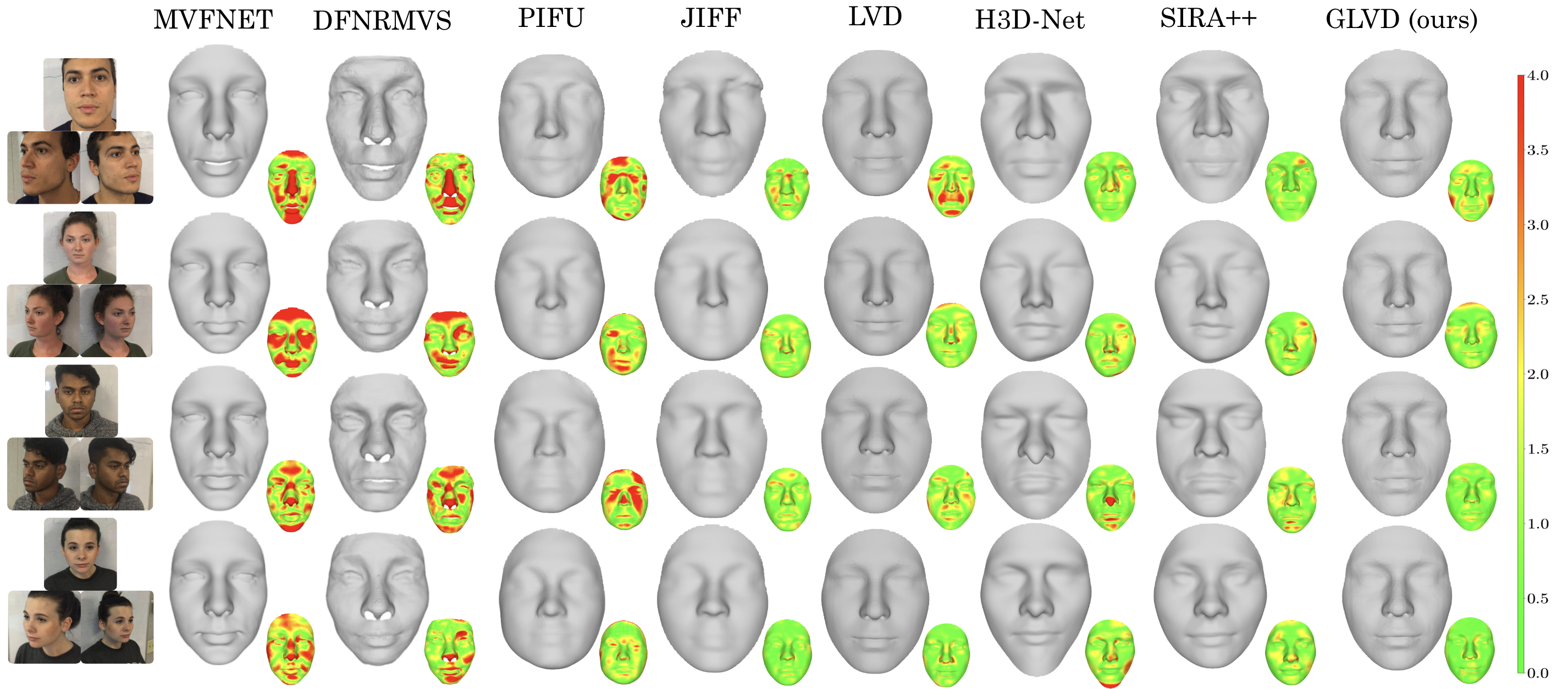}
    \caption{\textbf{Qualitative results on the 3DFAW dataset for three input images.} Each 3D reconstructed face is accompanied by a heatmap, where reddish
    areas indicate larger errors in mm.
    }
    \label{fig:qualitative_3view}
\end{figure}

\subsection{Vertex Displacement Prediction}

The vertex branch consists of two modules: a local feature extractor \( f_v : (\mathbf{I}, \mathbf{M}) \mapsto \mathbf{F}_{v} \) that computes image-aligned features for each projected vertex, and a regressor \( g_v(\cdot) \) that predicts the final vertex displacements \( \Delta \mathbf{v}_i \), which is also implemented as a 3-layer MLP.

We extract local image features \( \textbf{F}_{v} \) and \( \textbf{F}_k \) for each image \( \textbf{I} \), and following Sec.~\ref{sec:method_keypoints}, we estimate a set of predefined 3D keypoints at each iteration \( t \). 
Given a query vertex \( \mathbf{v}_i \in \mathbb{R}^3 \), we compute a keypoint-relative encoding matrix \( \mathbf{P} \in \mathbb{R}^{K \times 3} \), where each row represents a displacement vector \( \mathbf{P}^t = \mathbf{k}^t - \mathbf{v}_i^{t} \). As shown in our ablations (Sec.~\ref{sec:ablation}), this encoding outperforms alternatives such as euclidean distances, attention or concatenation.

\subsection{Displacement Learning and Optimization}

The network is trained to learn the parameters of \( f_v(\cdot) \), \( f_k(\cdot) \), \( g_k(\cdot) \), and \( g_v(\cdot) \). We use a dataset of \( N \) training scenes, each with a ground truth mesh with known topology, posed RGB images, and head masks. 
We sample query points for each scene using a hybrid strategy that combines uniform sampling with points near the mesh surface. Both model components are trained to predict displacements at each iteration \( t \). The iteration index is not explicitly encoded during training, as the model is exposed to a stochastic distribution of vertex states, making it inherently timestep-independent. At inference time, iteration \( t \) corresponds to updating each vertex by adding the displacement predicted at the previous step (Sec.~\ref{sec:lvd_background}). We pre-train the feature encoder on the 3D reconstruction task by augmenting it with a signed distance function (SDF) prediction head. This improves convergence behavior and leads to higher reconstruction accuracy. Further details are provided in the supplementary material.

\textbf{Keypoint Displacement Learning.} We select a consistent set of 3D keypoints to guide the surface displacement learning. For each mesh, we choose randomly a fixed subset of vertices to serve as target keypoints \( \{\mathbf{k}_j^t\} \). While their selection can be arbitrary, it must remain consistent across training scenes and during test-time inference. Each query keypoint, together with the input image \( \mathbf{I} \), is passed to the model \( (g_k \circ f_k)(x) \), which predicts 3D keypoint displacements \( \Delta\mathbf{k}_{i}^t \).

\textbf{Keypoint Displacement Learning.} We select a consistent set of 3D keypoints to guide the surface displacement learning. For each mesh, we choose the same subset of vertices to serve as target keypoints \( \{\mathbf{k}_j^t\} \). While their selection can be arbitrary, it must remain consistent across training scenes and during test-time inference. Each query keypoint, together with the input image \( \mathbf{I} \), is passed to the model \( (g_k \circ f_k)(x) \), which predicts 3D keypoint displacements \( \Delta\mathbf{k}_{i}^t \).

\input{Sections/05_main_table}

\textbf{Surface Displacement Learning.} We train \( f_v(\cdot) \) and \( g_v(\cdot) \) separately from the 3D keypoint branch. We encode vertices relative to the ground-truth 3D keypoints. To simulate prediction uncertainty, we perturb the sampled keypoints with noise drawn from a zero-mean multivariate Gaussian distribution with spherical covariance $\sigma^2$. These noisy keypoints are then used for encoding. Since depth errors in camera-aligned show higher variance, we apply noise with standard deviation \( 3\sigma \) along the depth axis in the camera frame.

Given a ground-truth mesh $\hat{V} = [\hat{\mathbf{v}}_{1}, \ldots, \hat{\mathbf{v}}_{N}]$ and its corresponding image $I$. We randomly sample $M$ 3D points $X = \{\mathbf{x}_{1}, \ldots, \mathbf{x}_{M}\}$, and compute the loss for each of the $M$ points:

\begin{equation}
    \mathcal{L} = \frac{1}{M} \sum_{i=1}^{M} \frac{1}{N} \sum_{j=1}^{N} 
    \left[ 
    \lambda_1 \left(1 - \frac{
        \Delta \mathbf{x}_i^j \cdot \hat{\Delta \mathbf{x}}_i^j
    }{
        \|\Delta \mathbf{x}_i^j\|_2 \, \|\hat{\Delta \mathbf{x}}_i^j\|_2
    }\right)
    + \lambda_2 \left| \|\Delta \mathbf{x}_i^j\|_2 - \|\hat{\Delta \mathbf{x}}_i^j\|_2 \right|
    \right]
\label{eq:loss}
\end{equation}

where \( \mathbf{x}_i \) is the \( i \)-th 3D query point, \( N \) is the number of ground-truth vertices, \( \Delta \mathbf{x}_i^j \) is the predicted displacement from \( \mathbf{x}_i \) to the \( j \)-th point, and \( \hat{\Delta \mathbf{x}}_i^j \) is the corresponding ground truth displacement. The symbol \( \cdot \) denotes the dot product, and \( \|\cdot\|_2 \) represents the Euclidean (L2) norm. The parameter \( \lambda_1 \) controls the contribution of the directional loss term, which minimizes the angular deviation \( 1 - \cos(\theta) \) between the predicted and ground truth vectors. The parameter \( \lambda_2 \) weights the magnitude loss, which penalizes differences in the length of the displacement vectors. To promote locality in the extracted image features, this term is clipped during training. This weighted combination encourages both directional consistency and training stability. During training, we apply binary dropout to the image features $\mathbf{F}_{v}$ to enhance robustness against unreliable predicted neural fields. Additionally, we model the 3D keypoints used for encoding vertices in the vertex branch as stochastic variables, introducing noise to the ground-truth keypoints only during training.

%% file: Sections/05_main_table.tex
\definecolor{rowblue}{RGB}{220,230,240}
\begin{table*}[t!]
\setlength{\tabcolsep}{4pt} 
\centering
\rowcolors{6}{rowblue}{white}
\caption{{\bfseries 3D face reconstruction comparison.} Average surface error (in mm) computed over all subjects in   3DFAW and H3DS datasets. We place "-" for not applicable configurations. Optimitzation-based have been included for reference.
}
\resizebox{0.79\textwidth}{!}{

\begin{tabular}{lcccccccccccc}
\toprule
& \multicolumn{4}{c}{3DFAW} & \multicolumn{8}{c}{H3DS 2.0} \\
\cmidrule(lr{0.5em}){2-5} \cmidrule(lr{0.5em}){6-13}
& \multicolumn{2}{c}{1 view} & \multicolumn{2}{c}{3 view} & \multicolumn{2}{c}{1 views} & \multicolumn{2}{c}{3 views} & \multicolumn{2}{c}{4 views} & \multicolumn{2}{c}{6 views}  \\
\cmidrule{1-13}
MVFNet \cite{wu2019mvf} & \multicolumn{2}{c}{-}& \multicolumn{2}{c}{1.56} & \multicolumn{2}{c}{-} & \multicolumn{2}{c}{1.73} & \multicolumn{2}{c}{-} & \multicolumn{2}{c}{-} \\
DFNRMVS \cite{bai2020deep} & \multicolumn{2}{c}{-}& \multicolumn{2}{c}{1.69}  & \multicolumn{2}{c}{-} & \multicolumn{2}{c}{1.83} & \multicolumn{2}{c}{-} & \multicolumn{2}{c}{-} \\
DECA \cite{feng2021learning} & \multicolumn{2}{c}{1.71}& \multicolumn{2}{c}{-}  & \multicolumn{2}{c}{1.99} & \multicolumn{2}{c}{-} & \multicolumn{2}{c}{-} & \multicolumn{2}{c}{-} \\
MICA \cite{MICA:ECCV2022} & \multicolumn{2}{c}{1.83}& \multicolumn{2}{c}{-}  & \multicolumn{2}{c}{2.08} & \multicolumn{2}{c}{-} & \multicolumn{2}{c}{-} & \multicolumn{2}{c}{-} \\
FaceVerse \cite{wang2022faceverse} & \multicolumn{2}{c}{1.88}& \multicolumn{2}{c}{-}  & \multicolumn{2}{c}{2.57} & \multicolumn{2}{c}{-} & \multicolumn{2}{c}{-} & \multicolumn{2}{c}{-} \\
FaceScape \cite{zhu2023facescape} & \multicolumn{2}{c}{1.61}& \multicolumn{2}{c}{-}  & \multicolumn{2}{c}{1.78} & \multicolumn{2}{c}{-} & \multicolumn{2}{c}{-} & \multicolumn{2}{c}{-} \\
HRN \cite{lei2023hierarchical}  & \multicolumn{2}{c}{1.60}& \multicolumn{2}{c}{-}  & \multicolumn{2}{c}{1.73} & \multicolumn{2}{c}{-} & \multicolumn{2}{c}{-} & \multicolumn{2}{c}{-} \\

VHAP \cite{qian2024vhap}  & \multicolumn{2}{c}{2.05}& \multicolumn{2}{c}{-}  & \multicolumn{2}{c}{2.15} & \multicolumn{2}{c}{-} & \multicolumn{2}{c}{-} & \multicolumn{2}{c}{-} \\

3DDFA-V3 \cite{wang20243d}  & \multicolumn{2}{c}{1.45}& \multicolumn{2}{c}{-}  & \multicolumn{2}{c}{1.65} & \multicolumn{2}{c}{-} & \multicolumn{2}{c}{-} & \multicolumn{2}{c}{-} \\
RAFaRe \cite{guo2023rafare}  & \multicolumn{2}{c}{1.68}& \multicolumn{2}{c}{-}  & \multicolumn{2}{c}{2.54} & \multicolumn{2}{c}{-} & \multicolumn{2}{c}{-} & \multicolumn{2}{c}{-} \\
PIFU \cite{saito2019pifu} & \multicolumn{2}{c}{2.19}& \multicolumn{2}{c}{1.99}  & \multicolumn{2}{c}{1.98} & \multicolumn{2}{c}{1.70} & \multicolumn{2}{c}{1.85} & \multicolumn{2}{c}{2.03}\\
JIFF \cite{cao2022jiff} & \multicolumn{2}{c}{1.48}& \multicolumn{2}{c}{1.47}  & \multicolumn{2}{c}{1.85} & \multicolumn{2}{c}{1.80} & \multicolumn{2}{c}{1.79} & \multicolumn{2}{c}{1.79}  \\
LVD~\citep{corona2022learned} & \multicolumn{2}{c}{1.58}& \multicolumn{2}{c}{1.26} & \multicolumn{2}{c}{1.39} & \multicolumn{2}{c}{1.45} & \multicolumn{2}{c}{1.39} & \multicolumn{2}{c}{1.37}  \\
\method (ours) & \multicolumn{2}{c}{\bfseries 1.25}& \multicolumn{2}{c}{\bfseries 1.22} & \multicolumn{2}{c}{\bfseries 1.36} & \multicolumn{2}{c}{\bfseries 1.33} & \multicolumn{2}{c}{\bfseries 1.34} & \multicolumn{2}{c}{\bfseries 1.34} \\
\cmidrule{1-13}
H3D-Net \cite{ramon2021h3d} & \multicolumn{2}{c}{1.70} & \multicolumn{2}{c}{1.37} & \multicolumn{2}{c}{-} & \multicolumn{2}{c}{1.44} & \multicolumn{2}{c}{1.41} & \multicolumn{2}{c}{1.21} \\
SIRA++~\citep{caselles2025implicit} & \multicolumn{2}{c}{1.35} & \multicolumn{2}{c}{1.32} & \multicolumn{2}{c}{1.57} & \multicolumn{2}{c}{1.18} & \multicolumn{2}{c}{1.23} & \multicolumn{2}{c}{1.04}\\

\bottomrule
\end{tabular}
}
\label{table:quantitative}
\end{table*}

%% file: Sections/05_experiments.tex
\section{Experiments}

\textbf{Training face dataset.} We employ a proprietary dataset of 3D head scans collected from 10,000 individuals, balanced by gender and diverse in age and ethnicity. All scans are aligned to a template 3D model using non-rigid Iterative Closest Point (ICP) registration for consistency.

\textbf{H3DS 2.0. \citep{ramon2021h3d, caselles2025implicit}} It contains 60 high-quality 3D full-head scans, including hair and shoulders, paired with posed RGB images. Each image includes a foreground mask and calibrated camera parameters.

\textbf{3DFAW. \citep{pillai20192nd}} This dataset provides videos recorded as well as mid-resolution 3D ground truth of the facial region. We select 5 male and 5 female scenes and use them to evaluate only the facial region.

\textbf{CelebA-HQ. \citep{Karras2018Progressive}} This dataset comprises 30k high-quality images at 1024×1024 resolution, derived from the original CelebA dataset. We selected a subset of 6 subjects for our qualitative evaluation.

\subsection{3D Face estimation}

We conducted a comprehensive comparison of our method with several 3DMM-based reconstruction works, including MVFNet~\cite{bai2020deep}, DFNRMVS~\cite{wu2019mvf}, DECA~\cite{feng2021learning}, MICA~\cite{MICA:ECCV2022}, FaceScape \cite{zhu2023facescape}, FaceVerse \cite{wang2022faceverse}, HRN \cite{lei2023hierarchical}, 3DDFA-v3~\citep{wang20243d} and VHAP~\cite{qian2024vhap}. Additionally, we compared our approach to the model-free methods PIFU~\cite{saito2019pifu}, JIFF~\cite{cao2022jiff}, RAFaRe\citep{guo2023rafare}, H3D-Net \cite{ramon2021h3d} SIRA++~\citep{caselles2025implicit} and hybrid method LVD~\citep{corona2022learned}. We used the unidirectional Chamfer distance for the quantitative evaluation, measuring the surface error from the ground truth to the predictions. The results of this comparison are summarized in Table \ref{table:quantitative}. Qualitative results for 3DFAW subjects are presented in Figure~\ref{fig:qualitative_1_view} for the single view and in Figure~\ref{fig:qualitative_3view} for the multiview setting. Results on H3DS are presented in Figure~\ref{fig:multiview}. We show in Figure \ref{fig:celeb} the estimated 3D face and the guiding keypoints.

\method demonstrates consistently strong performance across 3DFAW and H3DS evaluations. It leverages the structured topology of 3DMMs while explicitly addressing the constraint of shape representation to a predefined model space and the resulting bias toward average mean shape. As a result, \method outperforms 3DMM-based approaches, particularly in the single-view setting (Figure~\ref{fig:qualitative_1_view}). In comparison to model-free single forward pass methods such as PIFU, JIFF and RAFaRe, it demonstrates superior performance in surface reconstruction, yielding surfaces with reduced errors and a more realistic shape appearance.

\begin{figure}[!t]
    \centering
    \begin{minipage}[c]{0.50\textwidth} 
        \centering
        \includegraphics[width=\textwidth]{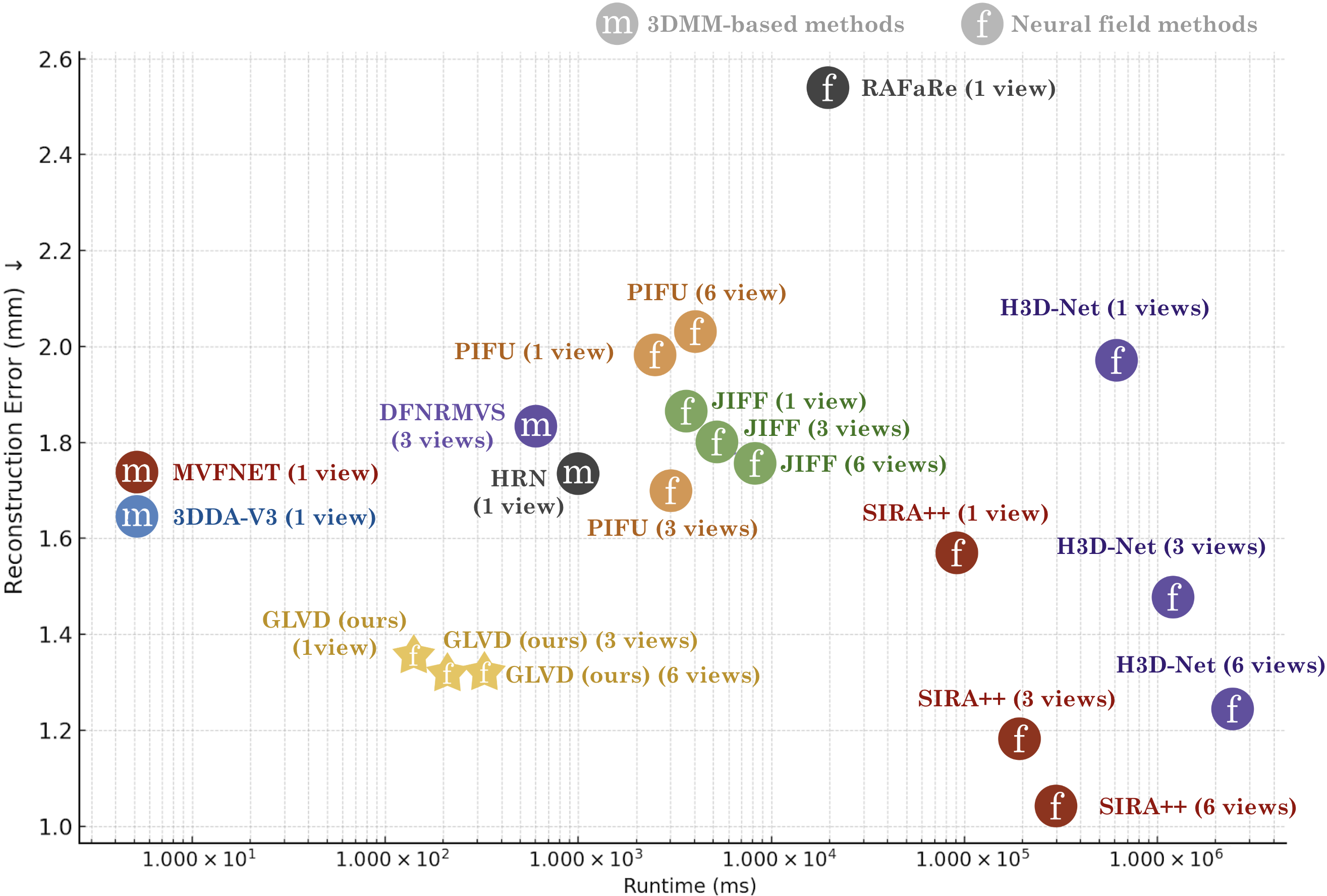}
    \end{minipage}%
    \hspace{0.10\textwidth} 
    \raisebox{0\height}{  
    \begin{minipage}[c]{0.35\textwidth} 
        \scriptsize
        \centering
        \renewcommand{\arraystretch}{1.2}
        \begin{tabular}[t]{ccc}
        \toprule
        & Test time & Test time \\
        & 1 view $\downarrow$ & 3 views $\downarrow$\\
        \cmidrule{1-3}
        
        MVFNet \citep{wu2019mvf} & \( - \) & \( <10ms \) \\
        DFNRMVS \citep{bai2020deep} & \( - \) & \( 0.6s \) \\
        3DDA-V3 \citep{wang20243d} & \( <10ms \) & \( - \) \\
        RAFaRe \citep{guo2023rafare} & \( 19s \) & \( - \) \\
        HRN \citep{hane2017hierarchical} & \( 1s \) & \( - \) \\
        PIFU \citep{saito2019pifu} & \( 2.5s \) & \( 3s \)\\
        JIFF \citep{cao2022jiff} & \( 4s \) & \( 5s \)\\
        H3D-Net \citep{ramon2021h3d} & \( 600s \) & \( 1200s \)\\
        SIRA++ \citep{Caselles_2023_SIRA} & \( 90s \) & \( 191s \)\\
        \method (ours) & \( 0.2s \) & \( 0.25s \) \\
        \bottomrule
        \end{tabular}
    \end{minipage}
    }
    \caption{\textbf{Quantitative comparison on H3DS dataset with one and three input views.} \textbf{Left.} Reconstruction error (mm) is plotted against runtime for various state-of-the-art methods under different view configurations. \textbf{Right.} Inference times for single and multi-view settings. 
    }
    \label{fig:time}
\end{figure}

Optimization-based methods are considered state-of-the-art for face reconstruction, particularly in multi-view settings where the problem becomes less ill-posed. However, their high computational cost remains a key limitation. In contrast, \method achieves comparable accuracy with over two orders of magnitude faster inference and without requiring postprocessing or template registration. Inference times are reported in Figure~\ref{fig:time} with 10 iterative update steps.

\subsection{Ablation}
\label{sec:ablation}

We conduct detailed ablation study on 3DFAW and H3DS 2.0 datasets to assess the impact of key design choices and demonstrate the effectiveness of the proposed method.

\begin{wraptable}{r}{0.4\textwidth} 
\centering
\captionsetup{font=small}
\caption{\textbf{Reconstruction quality comparison} using a single view with varying numbers of keypoints. Chamfer distance is reported in millimeters (mm).}
\begin{tabular}{l|cc}
\toprule
 & 3DFAW$\downarrow$ & H3DS$\downarrow$ \\
\midrule
\method- 4 & 1.92 & 1.51 \\
\method- 6 & 1.85 & 1.36 \\
\method- 12 & 1.57 & 1.42 \\
\method- 18 & \bfseries 1.25 & \bfseries 1.36 \\
\method- 24 & 1.58 & 1.33 \\
\bottomrule
\end{tabular}
\label{tab:lmks}
\end{wraptable}

\noindent
Table~\ref{tab:lmks} reports quantitative results using various subsets of 3D keypoints to guide \method. While the method is agnostic to landmark topology and supports flexible selection, we evaluate the impact of using different subsets derived from the template 3DMM. The specific landmark sets used are detailed in the supplementary material. Notably, results show that a small set of well-chosen keypoints provides effective global structural guidance, though limited coverage can introduce noise. While increasing the number of keypoints improves performance, gains in reconstruction accuracy diminish beyond a certain point. We attribute this to the fact that the keypoints are estimated. Higher-quality landmark supervision could further enhance reconstruction fidelity. In Figure~\ref{fig:celeb}, we provide face reconstruction results with the selected keypoints indicated for reference.

\textbf{Keypoints encoding.} In the original LVD formulation, global structure is not explicitly modeled, as point trajectories are predicted independently, with structural coherence learned implicitly through 2D feature volumes.
This results in ambiguity, as the model must resolve all possible correspondences per query without clear global guidance. In \method, we address this by introducing a landmark ensemble that serves as a compact global prior. These keypoints act as spatial anchors that guide vertex displacements, reducing ambiguity and promoting consistent topology. We ablate various encoding strategies in Table~\ref{tab:enc}. Replacing the per-vertex head with a global attention layer (k) introduces noise and instability, presumably due to long context. Inspired by the concept of skinning weights, we model vertex-to-keypoint relations via learnable attention (j), achieving performance on par with the standard encoding (l). Concatenation (h) and distance-based encoding (i) offer no gains, while removing absolute positions and using only relative encoding (g) leads to a performance drop.

\begin{wraptable}{r}{0.61\textwidth} 
\centering
\captionsetup{font=small}
\caption{Comparison of reconstruction quality for different number of keypoints. Chamfer distance in mm.}
\begin{tabular}{l|cc}
\toprule
 & 3DFAW$\downarrow$ & H3DS 2.0$\downarrow$\\
\midrule
(a) LVD  & 1.58 & 1.39 \\
(b) \method w/o loss (Eq.~\ref{eq:loss}) & 1.27 & 1.39 \\
(c) \method w/o HRNet & 1.52 & 1.39 \\
(d) \method w/o training Noise & 1.31 & 1.42 \\
(e) \method w/o binary dropout & 1.27 & 1.38 \\
(f) \method w/ canonical space & 1.51 & 1.52 \\
\midrule
(g) \method w/o vertex pos. & 1.55 & 1.67  \\
(h) \method w/ concat Enc. & 1.27 & 1.36 \\
(i) \method w/ norm Enc. & 1.26 & 1.37  \\
(j) \method w/ attention Enc. & 1.27 & 1.36  \\
(k) \method w/ attention layer & 1.85 & 1.97  \\
(l) \method  & \bfseries 1.25 & \bfseries 1.36  \\

\bottomrule
\end{tabular}
\label{tab:enc}
\end{wraptable}

\noindent
\textbf{Architecture analysis.} The \method architecture (Sec.~\ref{sec:method}) incorporates two core design choices: it estimates keypoints progressively during optimization and encodes vertices relative to these dynamically predicted keypoints. Table~\ref{tab:enc} presents an ablation of different training strategies. Using an L2 loss with clipping, following~\cite{corona2022learned} (b), results in unstable gradients and weak directional supervision. Incorporating a pre-trained HRNet for facial landmark prediction (c) highly improves performance in both vertex reconstruction and keypoint estimation. Injecting noise during training (d) to model the encoding as a stochastic variable improves robustness and test-time stability in the presence of uncertainty. Applying binary dropout to the 2D features during training (e) encourages reliance on the geometry-aware encoding and leads to better accuracy. Finally, optimizing in canonical space (f), as done in~\citep{ramon2021h3d, caselles2025implicit, gao2020portrait}, introduces a bias towards the mean shape. In the single-view setting, we use camera coordinates as the reference space, which removes the need for a calibrated camera at test time. We also investigate directly predicting the position of a specific vertex. We augment each query point with an identity feature derived from a Fourier embedding of the vertex’s 3D coordinates in the canonical face template. At test time, each sampled vertex is assigned a unique target identifier, and the model predicts displacements toward all possible targets, after which the displacement corresponding to the specified identifier is selected. However we empirically found it to be more stable and to yield better results by encoding trajectories implicitly as done in GLVD.

%% file: Sections/06_conclusion.tex
\begin{figure}[!t]
    \centering
    \includegraphics[width=0.95\textwidth]{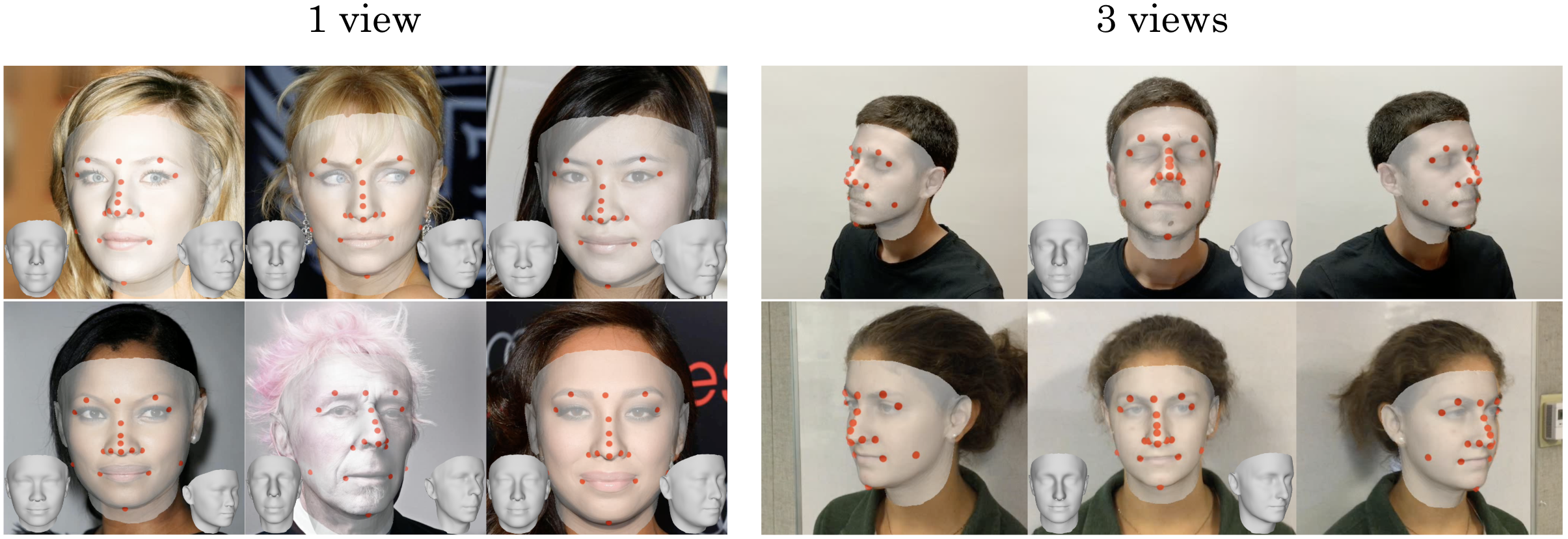}
    \caption{
    \textbf{Qualitative results for one and three input images.} Images are from CelebA-HQ (left), H3DS2.0 (right top), and 3DFAW (right bottom). At the last iteration step, we show the predicted template and the 3D keypoints in red.}
    \label{fig:celeb}
\end{figure}

\section{Discussion}

\textbf{Limitations and Future Work:} While GLVD demonstrates strong performance in few-shot 3D face reconstruction, it remain sensitive to occlusions and relies on the accuracy of keypoint predictions, which may degrade under challenging visual conditions. The focus of our method is the reconstruction of the face area by combining a hybrid method for fast and accurate prediction. Therefore, adding facial expressions is an interesting future direction. Future work may explore temporal consistency for video-based reconstruction and topology-adaptive strategies to better capture complex geometry.

\textbf{Conclusions.} In this paper, we have introduced \method, a hybrid approach for high-fidelity 3D face reconstruction from few-shot images. Our method introduces a novel combination of per-vertex neural fields and dynamically predicted 3D keypoints to provide both local accuracy and global structural guidance. By encoding vertex displacements relative to a sparse set of learned keypoints, our method refines mesh geometry iteratively without requiring parametric shape priors. The thorough evaluation demonstrated that our method achieves state-of-the-art performance in single-view settings and remains highly competitive in multi-view scenarios, all while substantially reducing inference time.

%% file: Sections/07_appendix.tex
\section{Appendix: Guided Learned Vertex Descend}
\label{sec:appendix}

In this appendix, we provide further technical details on
\begin{itemize}

    \item Experimental setup
    \item Different Keypoints configuration
    \item Implementation Details
    \item Additional qualitative results
    \item Failure Cases
    
\end{itemize}

For video results, including visual comparison to prior work, we refer to our supplementary video. This video includes a demonstration of \method for different input images.

\subsection{Experimental Setup}

\method adapts its reference space based on the number of input views during training and inference. For single-view 3D reconstruction, it operates in the camera coordinate frame, eliminating the need for camera parameter estimation at test time. In the multi-view setting, we canonicalize the 3D reconstruction and train a camera pose estimator on the same dataset used for training \method to enable prediction at inference time.

To ensure a fair comparison, PIFU, JIFF, LVD, and SIRA++ are trained using the same data used to train \method. While PIFU, LVD and JIFF were initially designed for full-body reconstruction, we modified their training to align with the data used by \method. To enhance robustness during training, we applied data augmentation techniques, including adjustments to brightness, contrast, hue, and saturation, as well as image jittering, blurring, and zooming. These augmentations are applied to the input images used for feature extraction. Additionally, we employed scene symmetrization, doubling the number of training scenes.

\subsection{Keypoints configuration}

\begin{figure}[h]
    \centering
    \includegraphics[width=1.0\textwidth]{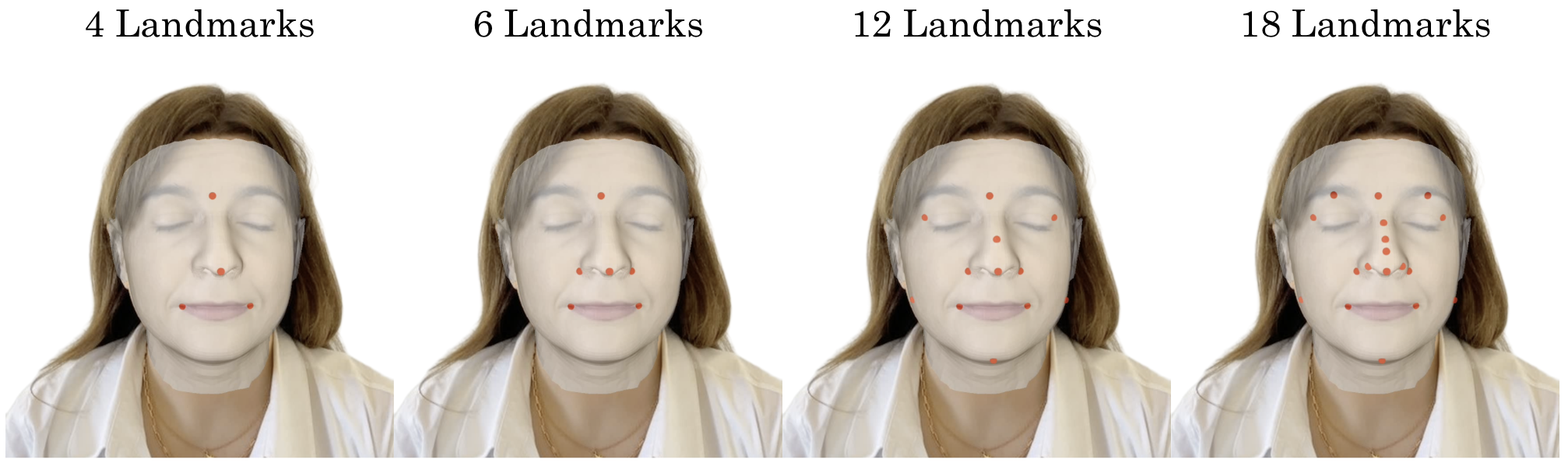}
    \caption{
    \textbf{Qualitative visualization of four keypoints configurations.} Images are from H3DS 2.0. At the last iteration step, we show the predicted template and the 3D keypoints in red.}
    \label{fig:sup_lmks}
\end{figure}

\method requires only  RGB images as input to predict the 3D surface. Internally, it operates by estimating 3D keypoints. Figure~\ref{fig:sup_lmks} presents visualizations of four proposed landmark subsets. The method is designed to allow a flexible selection of landmark configurations. In our experiments, we use template vertices registered to the training scenes. To adapt the method to other parametric models, such as FLAME or SMPL-X, joints can be selected as keypoints.

\subsection{Implementation Details}

The function \( f_v(\cdot) \) is a stacked hourglass network~\cite{newell2016stacked} composed of four stacks using group normalization~\cite{wu2018group}. Feature embeddings have a spatial resolution of \( 64 \times 64 \), with each containing \( 64 \) channels. As a result, each query point is represented by stacking four feature vectors of size \( 64 \times 4 = 256 \). We pre-train \( f_v(\cdot) \) to predict the signed distance function (SDF) values using the same training data. We set the clipping directional factor to 0.1, being the scene normalized in the centered cube of size 2.

The function \( f_k(\cdot) \) is implemented by a combination of facial keypoint heatmap estimator HRNet~\cite{wang2020deep} and a single-stack hourglass network~\cite{newell2016stacked}. During training, we keep the weights of the HRNet frozen. To generate \(  \textbf{F}_k \), we extract the first feature map computed with \( f_v(\cdot) \) and then concatenate it with the heatmaps predicted by HRNet. The combination of the features and the image \( \textbf{I} \) and the mask \( \textbf{M} \) is then fed to the hourglass network. Feature embeddings have a spatial resolution of \( 64 \times 64 \), with each containing \( 64 \) channels. We use a 0.1 clipping factor.

\begin{wraptable}{r}{0.4\textwidth} 
\centering
\captionsetup{font=small}
\caption{\textbf{Total number of parameters.}}
\begin{tabular}{l|cc}
\toprule
 & Parameters & Ratio \\
\midrule
\( f_v(\cdot) \) & 14.08 M & 54.9\% \\
\( g_v(\cdot) \) & 11.57 M & 45.1\% \\
Total & 25.64 M & 100\% \\
\midrule
\( f_k(\cdot) \) (HRNet) & 9.65 M & 67.2\% \\
\( f_s(\cdot) \) (Hourglass) & 4.28 M & 29.8\% \\
\( g_k(\cdot) \) & 0.43 M & 2.96\% \\
Total & 14.36 M & 100\% \\
\bottomrule
\end{tabular}
\label{tab:sup_model_size}
\end{wraptable}

Function \( g_v(\cdot) \) produces an output tensor of dimension $N = 7225 \times 3$. Given an input surface of size $7225 \times 3$, it outputs a tensor of shape $7225 \times 7225 \times 3$. Per-vertex displacements ($7225 \times 3$) are extracted from the diagonal and applied to update vertex positions.

\method works for different numbers of input images. When several images are used, we adopt a mean aggregation layer among features extracted from a multi-view feature encoder. In particular, we follow a single view forward pass independently of the number of input images until the second layer of the \( g_v(\cdot) \) and \( g_k(\cdot) \), where we apply a mean operation to aggregate multiview features.

Function \( g_k(\cdot) \) produces an output tensor of dimension $K = 18 \times 3$. Given an input surface of size $18 \times 3$, it outputs a tensor of shape $18 \times 18 \times 3$. Per-vertex displacements ($18 \times 3$) are extracted from the diagonal and applied to update keypoints positions. Both \( g_v(\cdot) \) and \( g_k(\cdot) \) are implemented as a 3-Layer MLP with ReLU activation and weight normalitzation.

\begin{figure}[!t]
    \centering
    \includegraphics[width=1.\textwidth]{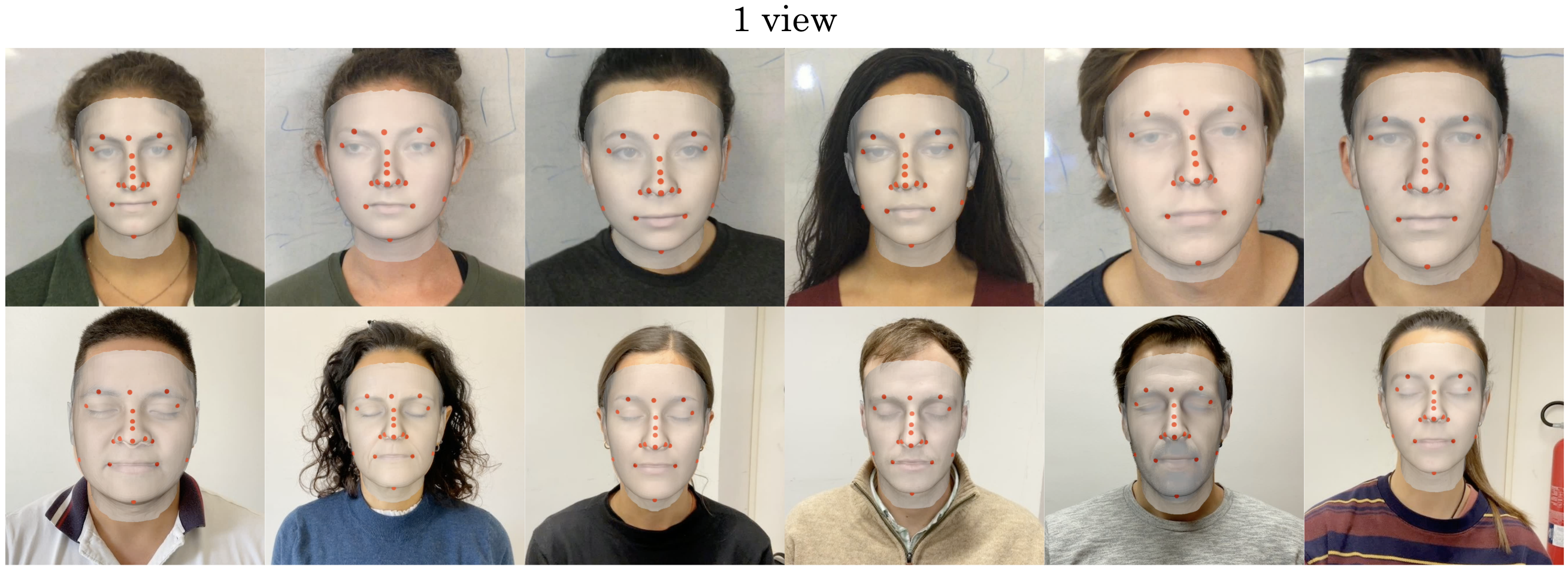}
    \caption{
    \textbf{Qualitative results for one input image.} Images are from 3DFAW (top), and H3DS 2.0 (bottom). At the last iteration step, we show the predicted template and the 3D keypoints in red.}
    \label{fig:sup_1_view}
\end{figure}

\begin{figure}[!t]
    \centering
    \includegraphics[width=1.\textwidth]{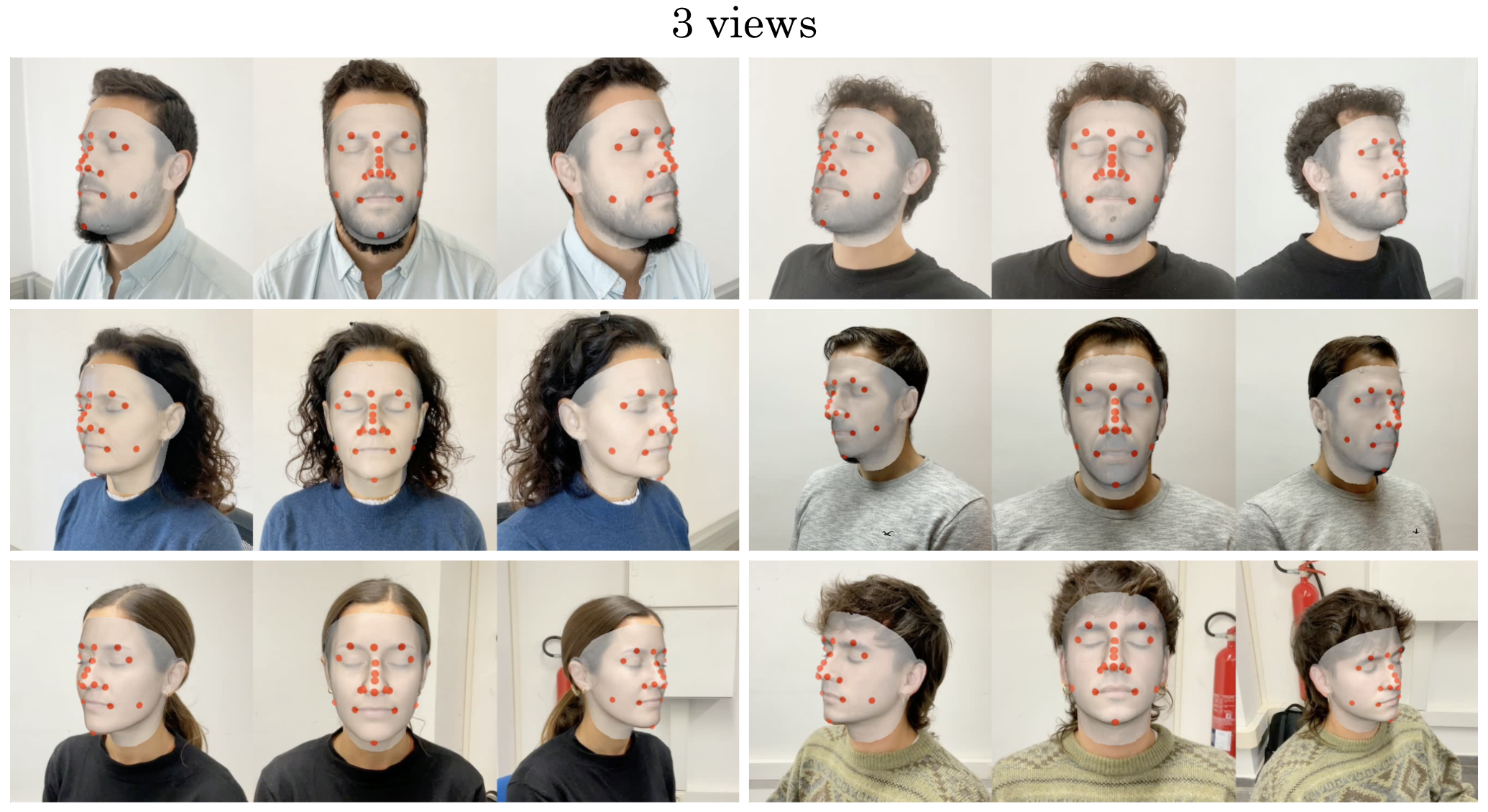}
    \caption{
    \textbf{Qualitative results for three input images.} Images are from 3DFAW. At the last iteration step, we show the predicted template and the 3D keypoints in red.}
    \label{fig:sup_3_views}
\end{figure}

All networks are trained end-to-end using GPU-accelerated hardware (RTX 4090). We use a batch size of 4 and an initial learning rate of 0.001 for 50 epochs, followed by 200 additional epochs with linear learning rate decay. For each scene, we sample 1400 vertices as query points. It takes between 1.5 to 6 days of training depending on the configuration. We set \( \lambda_1 = \lambda_2 = 0.5 \). Optimization is performed using Adam~\cite{kingma2014adam} with \(\beta_1 = 0.9\) and \(\beta_2 = 0.999\). The total number of parameters is detailed in Table~\ref{tab:sup_model_size}.

\subsection{Aditional results}

We provide qualitative results for subjects from 3DFAW and H3DS 2.0 using a single input view in Figure~\ref{fig:sup_1_view}, and for the multi-view setting in Figure~\ref{fig:sup_3_views}. We also show qualitative results in-the-wild CelebA-HQ dataset in Figure~\ref{fig:celeb}.

\begin{figure}[]
    \centering
    \includegraphics[width=0.95\textwidth]{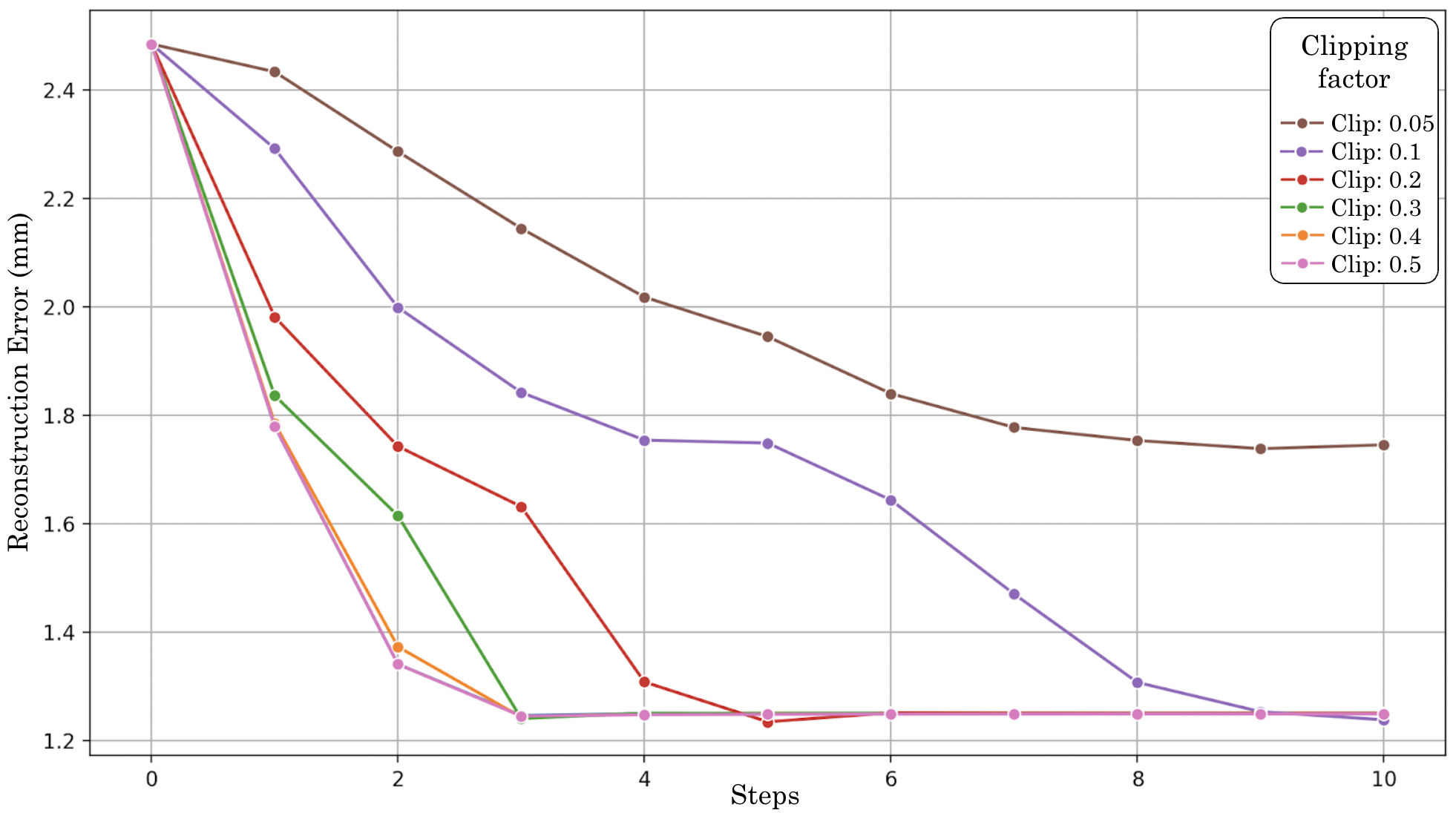}
    \caption{
    \textbf{Reconstruction Error from a Single Input Image.} Results report the mean Chamfer Distance on the 3DFAW dataset.
    }
    \label{fig:sup_fitting}
\end{figure}

Figure~\ref{fig:sup_fitting} reports the reconstruction error on 3DFAW subjects under varying numbers of update steps and clipping factor. At test time, the magnitude of the predicted displacement vector is clipped within the range [0.05, 0.5]. Results indicate that the number of update steps has a limited impact on reconstruction accuracy, which is strongly influenced by the clipping value used during training. This parameter controls the trade-off between accuracy and computational cost. Our experiments achieve the best performance with 10 steps and a clipping factor of 0.1.

We conducted an ablation study evaluating sequential vs parallel update strategies for the vertex and keypoints refinement modules on the H3DS 2.0 and 3DFAW datasets (Table \ref{tab:ablation_updates}). The results demonstrate that the iterative parallel update scheme yields consistently superior performance compared to the sequential alternative, although the improvement is minor. 

We also demonstrate that pre-training the feature encoder on a 3D reconstruction task, where it is trained to predict signed distance functions (SDFs), leads to faster convergence and improved performance. In this setting, (1) we pre-train the feature encoder on the 3D reconstruction task. We represent the surface $\mathcal{S}$ as the zero-level set of a signed distance function $f^{\text{sdf}} : (\mathbf{x}, I) \rightarrow s$, such that $\mathcal{S} = \{\mathbf{x} \in \mathbb{R}^3 \mid f^{\text{sdf}}(\mathbf{x}, I) = 0\}$. Our goal is to estimate $f^{\text{sdf}}$ through a composition of a feature encoder and a decoder network. The resulting feature encoder is then used within GLVD. To train on the SDF task, we use non-watertight scans from the same training dataset and minimize $\mathcal{L}_{\text{Surf}}^{(i)}$ on surface points $N_s$ and $\mathcal{L}_{\text{Eik}}^{(i)}$ throughout the volume $N_v$:

\begin{align}
    \mathcal{L}_{\text{Surf}} &= \frac{1}{N_s} \sum_{i=1}^{N_s} \big| f^{\text{sdf}}(\mathbf{x}_i, I) \big|, \\
    \mathcal{L}_{\text{Eik}}  &= \frac{1}{N_v} \sum_{i=1}^{N_v} 
    \left( \left\| \nabla_{\mathbf{x}} f^{\text{sdf}}(\mathbf{x}_i, I) \right\|_2 - 1 \right)^2 .
\end{align}

The loss is averaged over the batch. We compare in Table \ref{tab:sdf} the impact of pretraining.  We evaluate LVD with and without SDF-based pretraining of the feature encoder. Results show that this pretraining is crucial for achieving strong performance in both LVD and GLVD.

\begin{table}[t]
\centering
\caption{Ablation of sequential versus iterative (parallel) update strategies on H3DS 2.0 and 3DFAW.}

\label{tab:ablation_updates}
\begin{tabular}{lcccc}
\toprule
\textbf{H3DS 2.0 Dataset} & 1 view $\downarrow$ & 3 views $\downarrow$ & 4 views $\downarrow$ & 6 views $\downarrow$ \\
\midrule
LVD                   & 1.55 & 1.49 & 1.44 & 1.42 \\
LVD pre-trained SDF   & 1.39 & 1.45 & 1.39 & 1.37 \\
GLVD Sequential       & 1.38 & 1.36 & \textbf{1.34} & 1.35 \\
GLVD Iterative        & \textbf{1.36} & \textbf{1.33} & \textbf{1.34} & \textbf{1.34} \\
\midrule
\textbf{3DFAW Dataset} & 1 view $\downarrow$ & 3 views $\downarrow$ & \multicolumn{2}{c}{---} \\
\midrule
LVD                   & 1.65 & 1.52 & \multicolumn{2}{c}{---} \\
LVD pre-trained SDF   & 1.58 & 1.26 & \multicolumn{2}{c}{---} \\
GLVD Sequential       & 1.29 & 1.23 & \multicolumn{2}{c}{---} \\
GLVD Iterative        & \textbf{1.25} & \textbf{1.22} & \multicolumn{2}{c}{---} \\
\bottomrule
\end{tabular}
\label{tab:sdf}
\end{table}

\subsection{Failure Cases}

\begin{figure}[!h]
    \centering
    \includegraphics[width=0.9\textwidth]{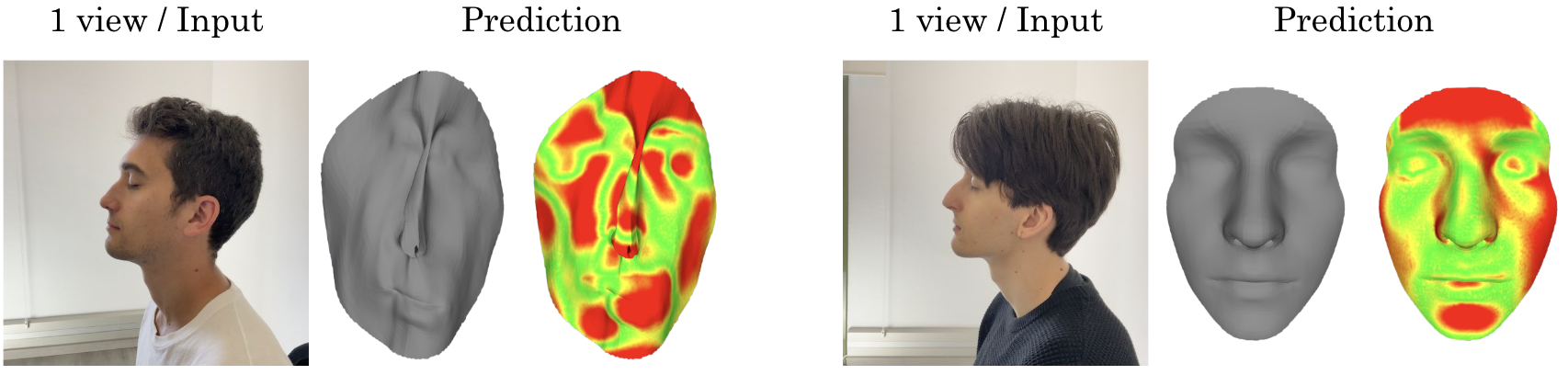}
    \caption{
    \textbf{Reconstruction Error from a Single Input Image.} Results report failure cases.}
    \label{fig:sup_fail}
\end{figure}

We present failure cases of \method under varying numbers of input views. Figure~\ref{fig:sup_fail} illustrates qualitative results in extreme scenarios, while Table~\ref{tab:failure} reports quantitative performance for different viewing angles in the single-view setting. The best performance is observed with front-facing input images. As the viewing angle increases, performance degrades significantly, primarily due to inaccuracies in landmark estimation under self-occlusion conditions.

\subsection{Social Impact}

GLVD advances 3D face modeling with high accuracy, enabling applications in graphics, AR/VR, and biometrics. However, it also raises concerns about privacy, surveillance, and identity misuse. High-fidelity face reconstruction can be used without consent or for impersonation, contributing to deepfake risks. To mitigate these issues, responsible deployment, fairness audits, and privacy safeguards are essential. While GLVD is a technical step forward, its societal implications must be carefully considered.

\begin{table}[h]
    \centering
    \caption{
    \textbf{Quantitative Results for a Single Input Image at varying input angles.} Chamfer Distance is reported in millimeters (mm) on the H3DS 2.0 dataset.
    }
    \begin{tabular}{l|ccc}
    \toprule
     & 1 view 0\textdegree\ & 1 view 45\textdegree\ & 1 view 90\textdegree\ \\
    \midrule
    \method & 1.31 & 2.07 & 2.11 \\

    \bottomrule
    \end{tabular}
    \label{tab:failure}
\end{table}

\begin{figure}[!h]
    \centering
    \includegraphics[width=1.0\textwidth]{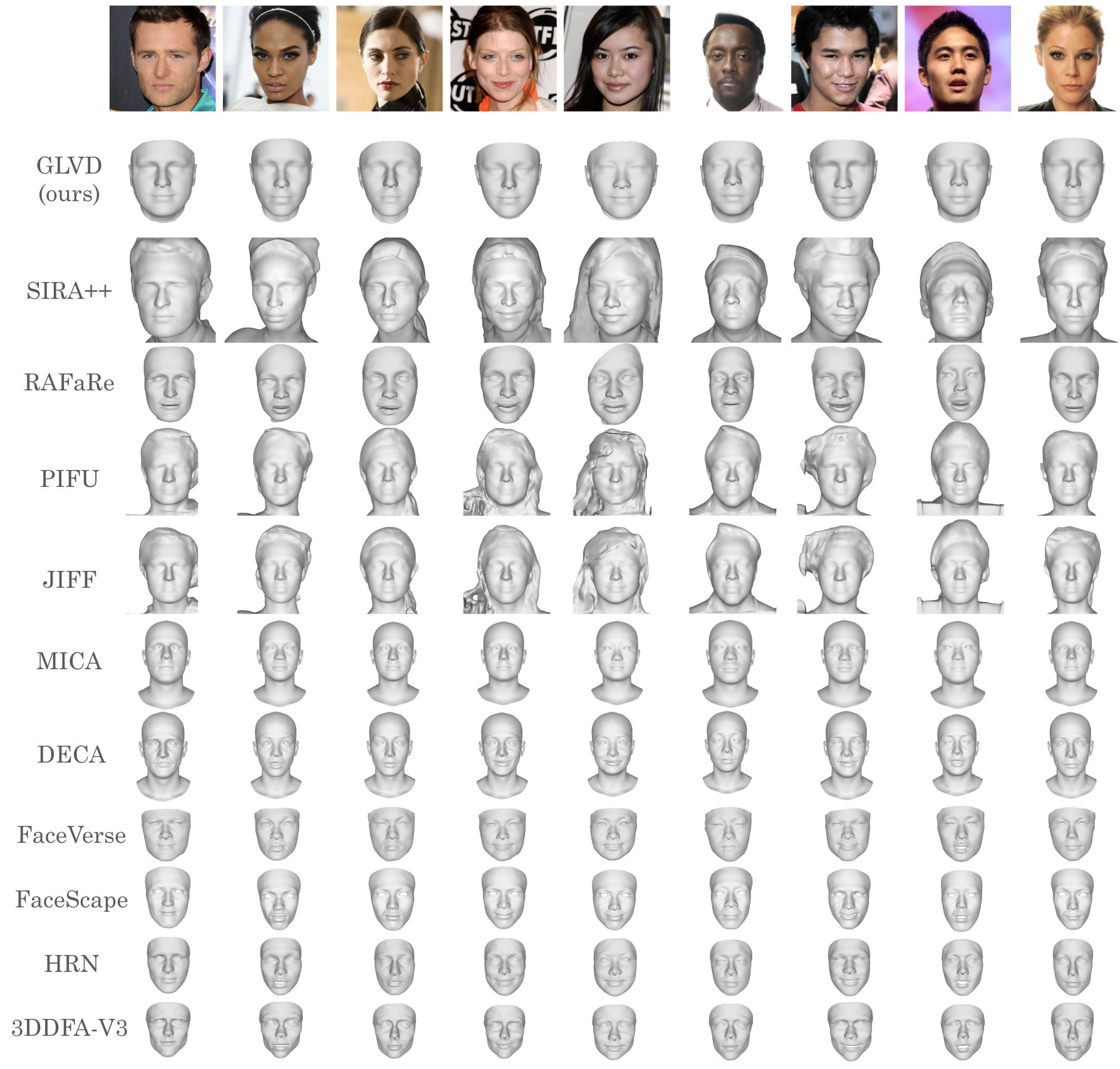}
    \caption{\textbf{Qualitative results on the CelebA-HQ dataset for a single input image.}
    }
    \label{fig:celeb_sup}
\end{figure}